\documentclass[11pt,a4paper]{article}
\usepackage[hyperref]{acl2020}
\usepackage{times}
\usepackage{latexsym}

\usepackage{times}  % DO NOT CHANGE THIS
\usepackage{helvet} % DO NOT CHANGE THIS
\usepackage{courier}  % DO NOT CHANGE THIS
\usepackage{graphicx} % DO NOT CHANGE THIS
\usepackage{color}
\usepackage{comment}
\usepackage{amsmath}
\usepackage{algorithm, algorithmic}
\usepackage{amssymb}
\usepackage{amsfonts}
\usepackage{mathtools}
\usepackage{multirow}
\usepackage{microtype}

\aclfinalcopy % Uncomment this line for the final submission
\title{Neural Topic Modeling with Bidirectional Adversarial Training}

\author{Rui Wang$^{\dag}$\ \ \ \  Xuemeng Hu$^{\dag}$\ \ \ \  Deyu Zhou$^{\dag}\thanks{corresponding author}$\ \ \ \ Yulan He$^{\S}$ \\ \textbf{Yuxuan Xiong}$^{\dag}$\ \ \ \   \textbf{Chenchen Ye}$^{\dag}$\ \ \ \ \textbf{Haiyang Xu}$^{\ddag}$\ \ \ \ \\
	$^{\dag}$School of Computer Science and Engineering, Key Laboratory of Computer Network\\
	and Information Integration, Ministry of Education, Southeast University, China \\
	$\S$Department of Computer Science, University of Warwick, UK \\
	$\ddag$AI Labs - Didi Chuxing Co., Ltd. - Beijing, China\\
	\{rui\_wang, xuemenghu, d.zhou, yuxuanxiong, chenchenye\}@seu.edu.cn,\\  yulan.he@warwick.ac.uk, xuhaiyangsnow@didiglobal.com}

\date{}

\begin{document}
\maketitle
\begin{abstract}
{\color{black}Recent years have witnessed a surge of interests of using neural topic models for automatic topic extraction from text, since they avoid the complicated mathematical derivations for model inference as in traditional topic models such as Latent Dirichlet Allocation (LDA). {\color{black}However, these models  either typically assume improper prior (e.g. Gaussian or Logistic Normal) over latent topic space or could not infer topic distribution for a given document.}
To address these limitations, 
we propose a neural topic modeling approach, called Bidirectional Adversarial Topic (BAT) model, which represents the first attempt of applying bidirectional adversarial training for neural topic modeling. The proposed BAT builds a two-way projection between the document-topic distribution and the document-word distribution. It uses a generator to capture the semantic patterns from texts and an encoder for topic inference. Furthermore, to incorporate word relatedness information, the Bidirectional Adversarial Topic model with Gaussian (Gaussian-BAT) is extended from BAT. To verify the effectiveness of BAT and Gaussian-BAT, three benchmark corpora are used in our experiments. The experimental results show that BAT and Gaussian-BAT obtain more coherent topics, outperforming several competitive baselines. Moreover, when performing text clustering based on the extracted topics,  
our models outperform all the baselines, with more significant improvements achieved by Gaussian-BAT where an increase of near 6\% is observed in accuracy.
}
\end{abstract}

\begin{comment}
{\color{blue}However, these models typically assume Gaussian prior or Logistic-Normal prior instead of Dirichlet prior which is more appropriate for topic modeling.}
\end{comment}

\section{Introduction}
Topic models have been extensively explored in the Natural Language Processing (NLP) community for unsupervised knowledge discovery. Latent Dirichlet Allocation (LDA)~\cite{blei2003latent}, the most popular topic model, has been extended~\cite{lin2009joint,zhou2014simple,cheng2014btm} for various extraction tasks. Due to the difficulty of exact inference, most LDA variants require approximate inference methods, such as mean-field methods and collapsed Gibbs sampling. However, these approximate approaches have the drawback that small changes to the modeling assumptions result in a re-derivation of the inference algorithm, which can be mathematically arduous. %and time-consuming

%\begin{comment}
%\end{comment}
\begin{figure}[t]%!htbp
\centering
\includegraphics[
  width=0.35\textwidth,
  keepaspectratio]
{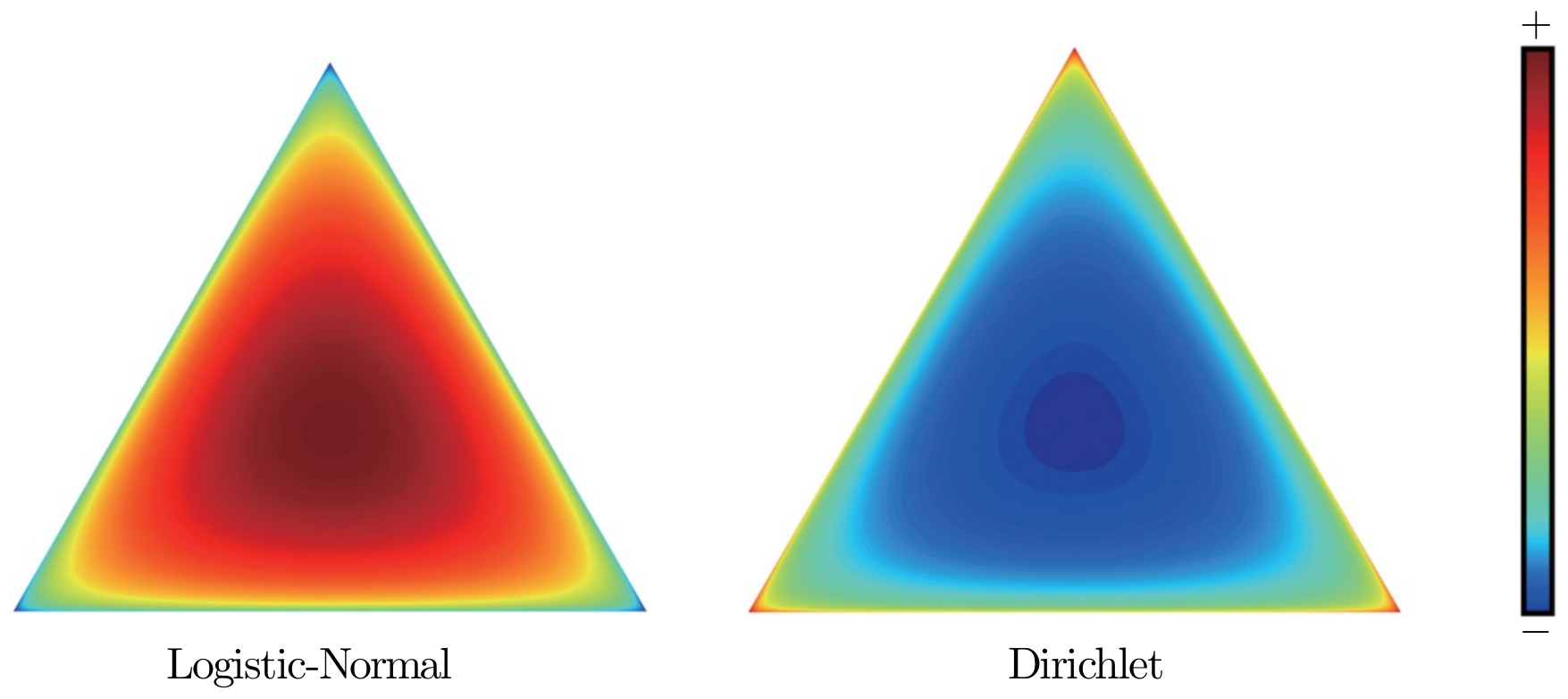}
\caption{Illustrated probability simplex with Logistic-Normal distribution and Dirichlet distribution. }
\label{fig:distributions}
\end{figure}

One possible way in addressing this limitation is through neural topic models which employ black-box inference mechanism with neural networks.  Inspired by variational autoencoder (VAE)~\cite{kingma2013auto},  Srivastava and Sutton~\shortcite{srivastava2017autoencoding} used the Logistic-Normal prior to mimic the simplex in latent topic space and proposed the Neural Variational LDA (NVLDA). Moreover, they replaced the word-level mixture in NVLDA with a weighted product of experts and proposed the ProdLDA~\cite{srivastava2017autoencoding} to further enhance the topic quality.

Although Srivastava and Sutton~\shortcite{srivastava2017autoencoding} used the Logistic-Normal distribution to approximate the Dirichlet distribution, they are not exactly the same. An illustration of these two distributions is shown in Figure~\ref{fig:distributions} in which the Logistic-Normal distribution does not exhibit multiple peaks at the vertices of the simplex as that in the Dirichlet distribution and as such, it is less capable to capture the multi-modality which is crucial in topic modeling~\cite{wallach2009rethinking}.  {\color{black}To deal with the limitation, Wang et al.~\shortcite{wang2019atm} proposed the Adversarial-neural Topic Model (ATM) based on adversarial training, it uses a generator network to capture the semantic patterns lying behind the documents. However, given a document, ATM is not able to infer the document-topic distribution which is useful for downstream applications, such as text clustering. }Moreover, ATM take the bag-of-words assumption and do not utilize any word relatedness information captured in word embeddings which have been proved to be crucial for better performance in many NLP tasks~\cite{liu2018content,lei2018saan}. 

{\color{black}To address these limitations, %shortages and ensure the model could provide topic-word distribution and document-topic distribution simultaneously, } 
we model topics with Dirichlet prior and propose a novel Bidirectional Adversarial Topic model (BAT) based on bidirectional adversarial training. The proposed BAT employs a generator network to learn the projection function from randomly-sampled document-topic distribution to document-word distribution. Moreover, an encoder network is used to learn the inverse projection, transforming a document-word distribution into a document-topic distribution.  Different from traditional models that often resort to analytic approximations, BAT employs a discriminator which aims to discriminate between real distribution pair and fake distribution pair, % as input} to recognize the real pairs from the fake ones 
thereby helps the networks (generator and encoder) to learn the two-way projections better.  During the adversarial training phase, the supervision signal provided by the discriminator will guide the generator to construct a more realistic document and thus better capture the semantic patterns in text. Meanwhile, the encoder network is also guided to generate a more reasonable topic distribution conditioned on specific document-word distributions. 
{\color{black}Finally, to incorporate the word relatedness information captured by word embeddings, we extend the BAT by modeling each topic with a multivariate Gaussian in the generator and propose the Bidirectional Adversarial Topic model with Gaussian (Gaussian-BAT). }}

The main contributions of the paper are:
\begin{itemize}
\item We propose a novel Bidirectional Adversarial Topic (BAT) model, which is, to our best knowledge, the first attempt of using bidirectional adversarial training in neural topic modeling;
\item We extend BAT to incorporate the word relatedness information into the modeling process and propose the Bidirectional Adversarial Topic model with Gaussian (Gaussian-BAT);
\item Experimental results on three public datasets show that BAT and Gaussian-BAT outperform the state-of-the-art approaches in terms of topic coherence measures. The effectiveness of BAT and Gaussian-BAT is further verified in text clustering.
\end{itemize}

\begin{figure*}[!h]
\centering
\includegraphics[
  width=0.9\textwidth,
  keepaspectratio]
{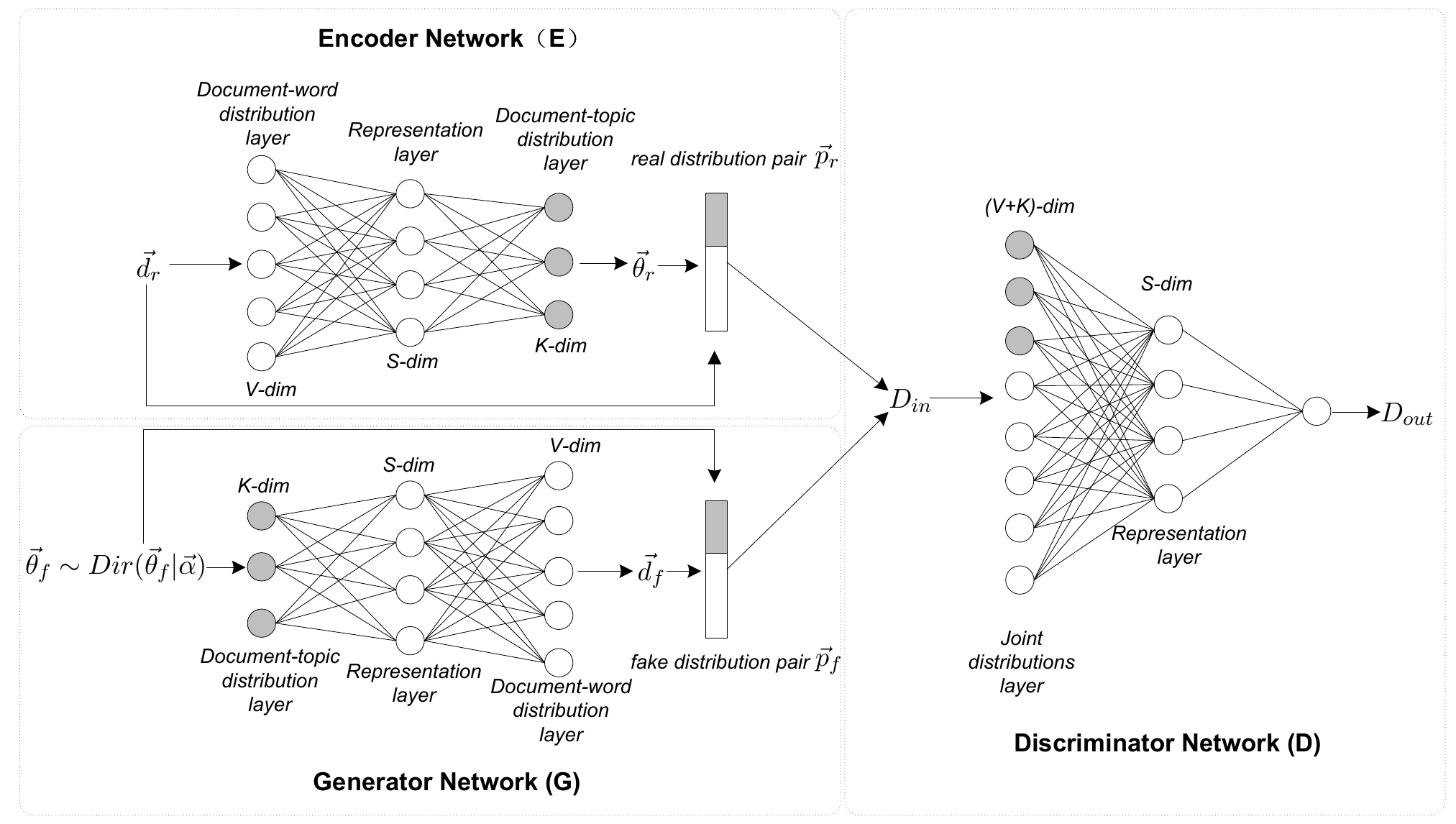}
\caption{The framework of the Bidirectional Adversarial Topic (BAT) model.}
\label{fig:bat_framework}
\end{figure*}

%{\color{black}Revised here...}
\section{Related work}
Our work is related to two lines of research, which are adversarial training and neural topic modeling.

\begin{comment}
\subsection{Word Representation Learning}
Distributed semantic models have recently been applied successfully in many NLP tasks~\cite{liu2018content,lei2018saan}. %Neural based models have been more efficient thanks to 
A notable example is \emph{word2vec} which relies on the skip-gram with negative sampling training method~\cite{mikolov2013efficient,mikolov2013distributed}. However, \emph{word2vec} only employs local context information. To address this limitation, Pennington~\shortcite{pennington2014glove} proposed a global log-bilinear regression model, called \emph{GloVe}, which combines advantages of the global matrix factorization and local context window methods. To deal with out-of-vocabulary words, \emph{FastText}~\cite{joulin2017bag} was proposed which generates word embeddings as the sum of their character-level $n$-grams representations. More recently, \emph{Probabilistic FastText}~\cite{athiwaratkun2018probabilistic} was proposed which is able to capture multiple word senses, sub-word structure, and uncertainty information jointly.
\end{comment}

\subsection{Adversarial Training}

Adversarial training, first employed in Generative Adversarial Network (GAN)~\cite{goodfellow2014generative}, has been extensively studied from both theoretical and practical perspectives.

Theoretically, Arjovsky~\shortcite{arjovsky2017wasserstein} and Gulrajani~\shortcite{gulrajani2017improved} proposed the Wasserstein GAN which employed the Wasserstein distance between data distribution and generated distribution as the training objective. To address the limitation that most GANs~\cite{goodfellow2014generative,radford2015unsupervised} could not project data into a latent space, Bidirectional Generative Adversarial Nets (BiGAN) ~\cite{donahue2016adversarial} and Adversarially Learned Inference (ALI)~\cite{dumoulin2016adversarially} were proposed.

%Moreover, various approaches based on 
Adversarial training has also been extensively used for text generation. For example, SeqGAN~\cite{yu2017seqgan} incorporated a policy gradient strategy for text generation. RankGAN~\cite{lin2017adversarial} ranked a collection of human-written sentences to capture the language structure for improving the quality of text generation. To avoid mode collapse when dealing with discrete data, MaskGAN~\cite{fedus2018maskgan} used an actor-critic conditional GAN to fill in missing text conditioned on the context.

\subsection{Neural Topic Modeling}

To overcome the challenging exact inference of topic models based on  directed graph, a replicated softmax model (RSM), based on the Restricted Boltzmann Machines was proposed in ~\cite{hinton2009replicated}. Inspired by VAE, Miao et al.~\shortcite{miao2016neural} used the multivariate Gaussian as the prior distribution of latent space and proposed the Neural Variational Document Model (NVDM) for text modeling. {\color{black}To model topic properly, the Gaussian Softmax Model (GSM)~\cite{miao2017discovering} which constructs the topic distribution using a Gaussian distribution followed by a softmax transformation was proposed based on the NVDM.} Likewise, to deal with the inappropriate Gaussian prior of NVDM, Srivastava and Sutton~\shortcite{srivastava2017autoencoding} proposed the NVLDA which approximates the Dirichlet prior using a Logistic-Normal distribution. %Furthermore, they also replaced the word-level mixture assumption used in NVLDA with a weighted product of experts~\cite{hinton2002training} and proposed the ProdLDA to enhance topic coherence. 
{\color{black} Recently, the Adversarial-neural Topic Model (ATM)~\cite{wang2019atm} is proposed based on adversarial training, it models topics with Dirichlet prior which is able to capture the multi-modality compared with logistic-normal prior and obtains better topics. Besides, the Adversarial-neural Event (AEM)~\cite{wang-etal-2019-open} model is also proposed for open event extraction by representing each event as an entity distribution, a location distribution, a keyword distribution and a date distribution.}

{\color{black}
Despite the extensive exploration of this research field, scarce work has been done to incorporate Dirichlet prior,  word embeddings and bidirectional adversarial training into neural topic modeling. In this paper, we propose two novel topic modeling approaches, called BAT and Gaussian-BAT, which are different from existing approaches in the following aspects: (1) Unlike NVDM, GSM, NVLDA and ProdLDA which model latent topic with Gaussian or logistic-normal prior, BAT and Gaussian-BAT explicitly employ Dirichlet prior to model topics; {\color{black}(2) Unlike ATM which could not infer topic distribution of a given document, BAT and Gaussian-BAT uses a encoder to generate the topic distribution corresponding to the document;} (3) Unlike neural topic models that only utilize word co-occurrence information, Gaussian-BAT models topic with multivariate Gaussian and incorporates the word relatedness into modeling process.
}

\section{Methodology}

Our proposed neural topic models are based on bidirectional adversarial training~\cite{donahue2016adversarial} and aim to learn the two-way non-linear projection between two high-dimensional distributions. In this section, we first introduce the Bidirectional Adversarial Topic (BAT) model
that only employs the word co-occurrence information. Then, built on BAT, we model topics with multivariate Gaussian in the generator of BAT and propose the Bidirectional Adversarial Topic model with Gaussian (Gaussian-BAT), which naturally incorporates word relatedness information captured in word embeddings into modeling process. 

\subsection{Bidirectional Adversarial Topic model}

As depicted in Figure~\ref{fig:bat_framework},  the proposed BAT consists of three components: (1) The \emph{Encoder} $E$ takes the $V$-dimensional document representation $\vec d_{r}$ sampled from text corpus $C$ as input and transforms it into the corresponding $K$-dimensional topic distribution $\vec \theta_{r}$; (2) The \emph{Generator} $G$ takes a random topic distribution $\vec \theta_{f}$ drawn from a Dirichlet prior as input and generates a $V$-dimensional fake word distribution $\vec d_{f}$; (3) The \emph{Discriminator} $D$ takes the real distribution pair $\vec p_{r}=[\vec \theta_{r};\vec d_{r}]$ and fake distribution pair $\vec p_{f}=[\vec \theta_{f};\vec d_{f}]$ as input and discriminates the real distribution pairs from the fake ones. The outputs of the discriminator are used as supervision signals to learn $E$, $G$ and $D$ during adversarial training.  In what follows, we describe each component in more details.
%We will give a more detailed illustration of these components.  

\subsubsection{Encoder Network}$\\$
The encoder learns a mapping function to transform document-word distribution to document-topic distribution. As shown in the top-left panel of Figure~\ref{fig:bat_framework}, it contains a $V$-dimensional document-word distribution layer, an $S$-dimensional representation layer and a $K$-dimensional document-topic distribution layer, where $V$ and $K$ denote vocabulary size and topic number respectively. 

More concretely, for each document $d$ in text corpus, $E$ takes the document representation $\vec d_{r}$ as input, where $\vec d_{r}$ is the representation weighted by TF-IDF, and it is calculated by:
\begin{align}
tf_{i,d}&=\frac{n_{i,d}}{\sum_{v}n_{v,d}},\quad \quad  idf_{i}=\log \frac{|C|}{|C_{i}|} \notag \\
tf\textrm{-}idf_{i,d}&=tf_{i,d}* idf_{i}, \quad\ \ \   d_{r}^{i}=\frac{tf\textrm{-}idf_{i,d}}{\sum_{v}tf\textrm{-}idf_{v,d}}\notag
\end{align}
where $n_{i,d}$ denotes the number of $i$-th word appeared in document $d$, $|C|$ represents the number of documents in the corpus, and $|C_{i}|$ means the number of documents that contain $i$-th word in the corpus. Thus, each document could be represented as a $V$-dimensional multinomial distribution and the $i$-th dimension denotes the semantic consistency between $i$-th word and the document. 

With $\vec d_{r}$ as input, $E$ firstly projects it into an $S$-dimensional semantic space through the representation layer as follows:
\begin{align}
\vec h_{s}^{e}&= {\rm BN}(W_{s}^{e}\vec d_{r}+\vec b_{s}^{e})\label{eqs:encoder_s}\\
\vec o_{s}^{e}=&\max (\vec h_{s}^{e}, leak* \vec h_{s}^{e})
\end{align}
where $W_{s}^{e}\in \mathbb{R}^{S\times V}$ and $\vec b_{s}^{e}$ are weight matrix and bias term of the representation layer, $\vec h_{s}^{e}$ is the state vector normalized by batch normalization ${\rm BN}(\cdot)$,   $leak$ denotes the parameter of LeakyReLU activation and $\vec o_{s}^{e}$ represents the output of representation layer.  

Then,  the encoder transforms $\vec o_{s}^{e}$ into a $K$-dimensional topic space based on the equation below: 
\begin{equation}
\vec \theta_{r}={\rm softmax}(W_{t}^{e}\vec o_{s}^{e}+\vec b_{t}^{e})
\end{equation}
where $W_{t}^{e}\in \mathbb{R}^{K\times S}$ is the weight matrix of topic distribution layer, $\vec b_{t}^{e}$ represents the bias term, $\vec \theta_{r}$ denotes the corresponding topic distribution of the input $\vec d_{r}$ and the $k$-th ($k \in \{1,2,...,K\}$) dimension $\theta_{r}^{k}$ represents the proportion of $k$-th topic in document $d$.

\subsubsection{Generator network}$\\$
The generator $G$ is shown in the bottom-left panel of Figure~\ref{fig:bat_framework}. Contrary to encoder, it provides an inverse projection from document-topic distribution to document-word distribution and contains a $K$-dimensional document-topic layer, an $S$-dimensional representation layer and a $V$-dimensional document-word distribution layer.

As pointed out in~\cite{wallach2009rethinking}, the choice of Dirichlet prior over topic distribution is important to obtain interpretable topics. Thus, BAT employs the Dirichlet prior parameterized with $\vec \alpha$ to mimic the multi-variate simplex over topic distribution $\vec \theta_{f}$. It can be drawn randomly based on the equation below:
\begin{comment}
\begin{equation}
p(\vec \theta_{f}|\vec \alpha)=Dir(\vec \theta_{f}|\vec \alpha)\triangleq \frac{1}{\Delta(\vec \alpha)}\prod_{k=1}^{K}\left[\theta_{f}^{k}\right]^{\alpha_{k}-1}
\end{equation}
\end{comment}
\begin{equation}
p(\vec \theta_{f}|\vec \alpha)=Dir(\vec \theta_{f}|\vec \alpha)\triangleq \frac{1}{\Delta(\vec \alpha)}\prod_{k=1}^{K}\left[\theta_{f}^{k}\right]^{\alpha_{k}-1}
\end{equation}
where $\vec \alpha$ is the $K$-dimensional hyper-parameter of Dirichlet prior, $K$ is the topic number that should be set in BAT, $\theta_{f}^{k}\in \left[0,1\right]$, follows the constrain that $\sum_{k=1}^{K}\theta_{f}^{k}=1$, represents the proportion of the $k$-th topic in the document, and normalization term $\Delta(\vec \alpha)$ is defined as $\frac{\prod_{k=1}^{K}\Gamma(\alpha_{k})}{\Gamma(\sum_{k=1}^{K}\alpha_{k})}$.

To learn the transformation from document-topic distribution to document-word distribution, $G$ firstly projects $\vec \theta_{f}$ into an $S$-dimensional representation space based on equations:
\begin{align}
\vec h_{s}^{g}&={\rm BN}(W_{s}^{g}\vec \theta_{f}+\vec b_{s}^{g})\\
\vec o_{s}^{g}&=\max (\vec h_{s}^{g}, leak*\vec h_{s}^{g})\label{eqs:generator}
\end{align}
where $W_{s}^{g}\in \mathbb{R}^{S\times K}$ is weight matrix of the representation layer, $\vec b_{s}^{g}$ represents bias term, $\vec h_{s}^{g}$ is the state vector normalized by batch normalization, Eq.~\ref{eqs:generator} represents the LeakyReLU activation parameterized with $leak$, and $\vec o_{s}^{g}$ is the output of the representation layer.

{\color{black}Then, to project $\vec o_{s}^{g}$ into word distribution $\vec d_{f}$, a subnet contains a linear layer and a softmax layer is used and the transformation follows:
\begin{equation}
\vec d_{f}={\rm softmax}(W_{w}^{g}\vec o_{s}^{g}+\vec b_{w}^{g})\\
\end{equation}
where $W_{w}^{g}\in \mathbb{R}^{V\times S}$ and $\vec b_{w}^{g}$ are weight matrix and bias of word distribution layer, $\vec d_{f}$ is the word distribution correspond to $\vec \theta_{f}$. For each $v\in\{1,2,...,V\}$, the $v$-th dimension $d_{f}^{v}$ is the probability of the $v$-th word in fake document $\vec d_{f}$.
}

\subsubsection{Discriminator network}$\\$
The discriminator $D$ is constituted by three layers (a $V+K$-dimensional joint distribution layer, an $S$-dimensional representation layer and an output layer) as shown in the right panel of Figure~\ref{fig:bat_framework}. It employs real distribution pair $\vec p_{r}$ and fake distribution pair $\vec p_{f}$ as input and then outputs $D_{out}$ to identify the input sources (fake or real). Concretely, a higher value of $D_{out}$ represents that $D$ is more prone to predict the input as real and vice versa.

\subsection{BAT with Gaussian (Gaussian-BAT)}

\begin{figure}[!h]
\centering
\includegraphics[
  width=0.48\textwidth,
  keepaspectratio]
{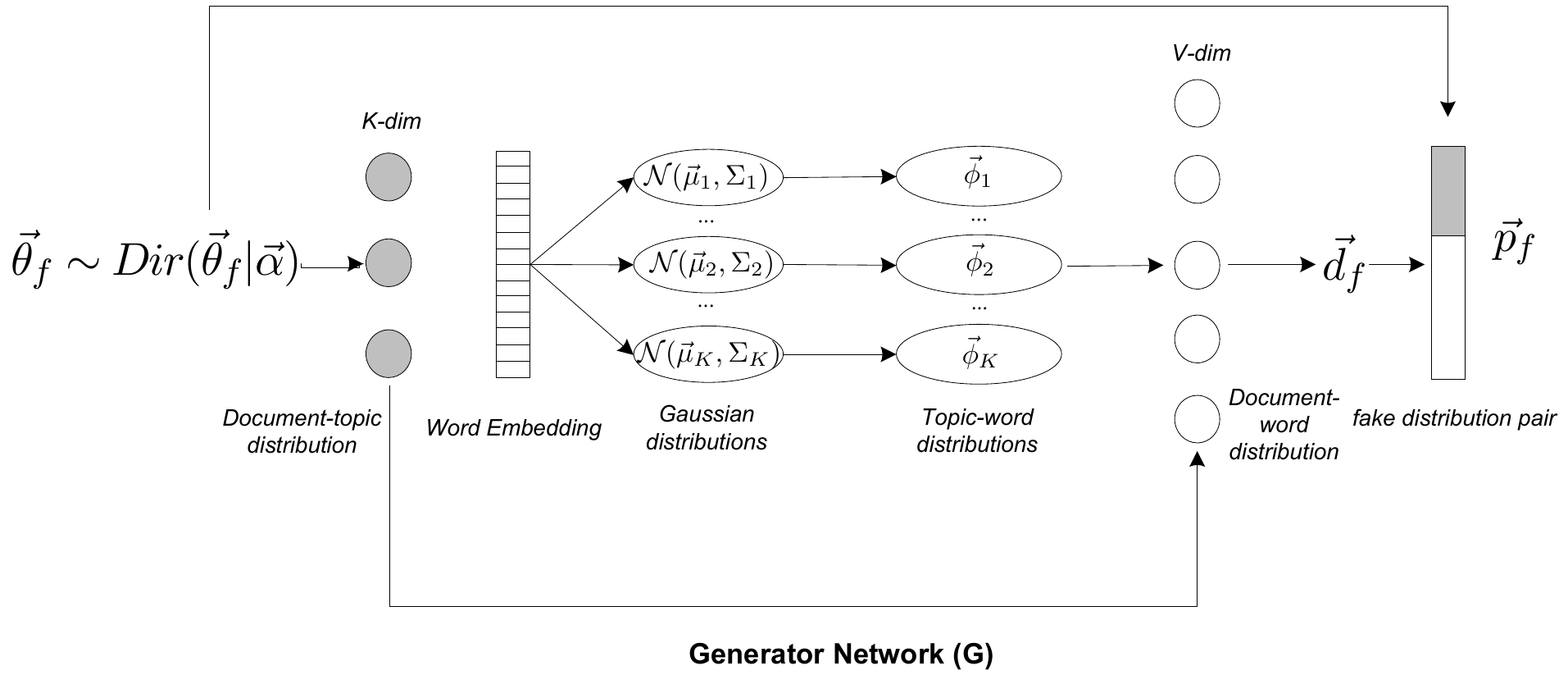}
\caption{The generator of Gaussian-BAT.}
\label{fig:generator_gaussian}
\end{figure}

In BAT, the generator models topics based on the bag-of-words assumption as in most other neural topic models. %and do not use incorporate word relatedness which has been proved to be helpful for many NLP tasks.
To incorporate the word relatedness information captured in word embeddings~\cite{mikolov2013efficient,mikolov2013distributed,pennington2014glove,joulin2017bag,athiwaratkun2018probabilistic} into the inference process, we modify the generator of BAT and propose Gaussian-BAT, in which $G$ models each topic with a multivariate Gaussian as shown in {\color{black}Figure~\ref{fig:generator_gaussian}}. 

Concretely, Gaussian-BAT employs the multivariate Gaussian $\mathcal N(\vec \mu_{k},\Sigma_{k})$ to model the $k$-th topic. Here, $\vec \mu_{k}$ and $\Sigma_{k}$ are trainable parameters, they represent mean and covariance matrix respectively. Following its probability density, for each word $v\in \{1,2,...,V\}$, the probability in the $k$-th topic $\phi_{k,v}$ is calculated by:
\begin{align}
p(\vec e_{v}|&topic=k)=\mathcal N(\vec e_{v};\vec \mu_{k},\Sigma_{k})\notag \\
&=\frac{\exp(-\frac{1}{2}(\vec e_{v}-\vec \mu_{k})^{\text{T}}\Sigma_{k}^{-1}(\vec e_{v}-\vec \mu_{k}))}{\sqrt{(2\pi)^{D_{e}}|\Sigma_{k}|} }\\
\phi_{k,v}&=\frac{p(\vec e_{v}|topic =k)}{\sum_{v=1}^{V}p(\vec e_{v}|topic=k)}
\end{align}
where $\vec e_{v}$ means the word embedding of $v$-th word, $V$ is the vocabulary size, $|\Sigma_{k}|=\det\ \Sigma_{k}$ is the determinant of covariance matrix $\Sigma_{k}$, $D_{e}$ is the dimension of word embeddings, $p(\vec e_{v}|topic=k)$ is the probability calculated by density, and $\vec \phi_{k}$ is the normalized word distribution of $k$-th topic. 
With randomly sampled topic distribution $\vec \theta_{f}$ and the calculated topic-word distributions $\{\vec \phi_{1},\vec \phi_{2},...,\vec \phi_{K}\}$, the fake word distribution $\vec d_{f}$ corresponding to $\vec \theta_{f}$ can be obtained by: 
\begin{equation}
\vec d_{f}=\sum_{k=1}^{K}\vec \phi_{k}*\theta_{k}
\end{equation}
where $\theta_{k}$ is the topic proportion of the $k$-th topic. Then, $\vec \theta_{f}$ and $\vec d_{f}$ are concatenated to form the fake distribution pair $\vec p_{f}$ as shown in Figure~\ref{fig:generator_gaussian}. And encoder and discriminator of Gaussian-BAT are same as BAT, shown as Figure~\ref{fig:bat_framework}. In our experiments, the pre-trained 300-dimensional \emph{Glove}~\cite{pennington2014glove} embedding is used.

\subsection{Objective and Training Procedure}

In Figure~\ref{fig:bat_framework}, the real distribution pair $\vec p_{r}=[\vec \theta_{r};\vec d_{r}]$ and the fake distribution pair $\vec p_{f}=[\vec \theta_{f};\vec d_{f}]$ can be viewed as random samples drawn from two $(K+V)$-dimensional joint distributions $\mathbb{P}_{r}$ and $\mathbb{P}_{f}$, each of them comprising of a $K$-dimensional Dirichlet distribution and a $V$-dimensional Dirichlet distribution.  The training objective of BAT and Gaussian-BAT is to make the generated joint distribution $\mathbb{P}_{f}$ close to the real joint distribution $\mathbb{P}_{r}$ as much as possible. %After model learning, the marginal distributions $\mathbb{P}_{r}^{\theta}$, $\mathbb{P}_{f}^{\theta}$, $\mathbb{P}_{r}^{d}$ and $\mathbb{P}_{f}^{d}$ have the following relations:
%\begin{align}
%\mathbb{P}_{r}^{\theta}=\mathbb{P}_{f}^{\theta},\quad\mathbb{P}_{f}^{d}=\mathbb{P}_{r}^{d}
%\end{align}
%where $\mathbb{P}_{r}^{\theta}$ and $\mathbb{P}_{f}^{\theta}$ are $K$-dimensional Dirichlet distributions of $\vec \theta_{r}$ and $\vec \theta_{f}$, $\mathbb{P}_{r}^{d}$ and $\mathbb{P}_{f}^{d}$ are $V$-dimensional Dirichlet distributions of $\vec d_{r}$ and $\vec d_{f}$. 
In this way, a two-way projection between document-topic distribution and document-word distribution could be built by the learned encoder and generator.

%Due to the choice of distance which measures the distance between $\mathbb{P}_{r}$ and $\mathbb{P}_{f}$ is crucial for adversarial training, to deal with the limitation that Jensen-Shannon divergence ~\cite{goodfellow2014generative} that do not continuous with respect to the generator's parameters, 
To measure the distance between $\mathbb{P}_{r}$ and $\mathbb{P}_{f}$, we use the Wasserstein-distance as the optimization objective, since it was shown to be more effective compared to Jensen-Shannon divergence~\cite{arjovsky2017wasserstein}: %which follows the definition:
\begin{align}
Loss=\mathbb{E}_{\vec p_{f}\sim \mathbb{P}_{f}}\left[D(\vec p_{f})\right]-\mathbb{E}_{\vec p_{r}\sim \mathbb{P}_{r}}\left[D(\vec p_{r})\right]
\end{align}
where $D(\cdot)$ represents the output signal of the discriminator. A higher value denotes that the discriminator is more prone to consider the input as a real distribution pair and vice versa. In addition, we use weight clipping which was proposed to ensure the Lipschitz continuity~\cite{arjovsky2017wasserstein} of $D$.% 
\begin{algorithm}[!h]
	\renewcommand{\algorithmicrequire}{\textbf{Input:}}
	\renewcommand{\algorithmicensure}{\textbf{Output:}}
	%\vspace{-10pt}
	\caption{Training procedure for BAT and Gaussian-BAT}
	\label{alg:1}
	\begin{algorithmic}[1]
		\REQUIRE $K$, $c$, $n_{d}$, $m$, $\alpha_{1}$, $\beta{1}$, $\beta{2}$
		\ENSURE The trained encoder $E$ and generator $G$.
		\STATE Initialize $D$, $E$ and $G$ with $\omega_{d}$, $\omega_{e}$ and $\omega_{g}$
		\WHILE{ $\omega_{e}$ and $\omega_{g}$ have not converged}
		\FOR{$t=1,...,n_{d}$}
		\FOR{$j=1,...,m$}
		\STATE Sample $\vec d_{r}\sim \mathbb{P}_{r}^{d}$, 
		\STATE Sample a random  $\vec\theta_{f}\sim Dir(\vec\theta_{f}|\vec\alpha)$ 
		\STATE $\vec d_{f}\leftarrow G(\vec\theta_{f})$, $\vec \theta_{r}\leftarrow E(\vec d_{r})$
		\STATE $\vec p_{r}=[\vec \theta_{r};\vec d_{r}]$, $\vec p_{f}=[\vec \theta_{f};\vec d_{f}]$
		\STATE {\color{black}$L^{(j)}=D(\vec p_{f})-D(\vec p_{r})$}
		\ENDFOR
		\STATE $\omega_{d}\leftarrow Adam(\nabla_{\omega_{d}}\frac{1}{m}\sum_{j=1}^{m}L^{(j)},\omega_{d},p_{a}) $
		\STATE {\color{black}$\omega_{d}\leftarrow {\rm clip}(\omega_{d},-c,c)$}
		\ENDFOR
		\STATE $\omega_{g}\leftarrow Adam(\nabla_{\omega_{g}}\frac{-1}{m}\sum_{j=1}^{m}D(\vec p_{f}^{j}),\omega_{g},p_{a})$
		\STATE $\omega_{e}\leftarrow Adam(\nabla_{\omega_{e}}\frac{1}{m}\sum_{j=1}^{m}D(\vec p_{r}^{j}),\omega_{e},p_{a})$
		\ENDWHILE 	
	\end{algorithmic}
\end{algorithm}

The training procedure of BAT and Gaussian-BAT is given in Algorithm.~\ref{alg:1}. Here, $c$ is the clipping parameter, $n_{d}$ represents the number of discriminator iterations per generator iteration, $m$ is the batch size, $\alpha_{1}$ is the learning rate, $\beta_{1}$ and $\beta_{2}$ are hyper-parameters of Adam~\cite{kingma2014adam}, and $p_{a}$ represents $\{\alpha_{1},\beta_{1},\beta_{2}\}$. In our experiments, we set the $n_{d}=5$, $m=64$, $\alpha_{1}=1e-4$, $c=0.01$, $\beta_{1}=0.5$ and $\beta_{2}=0.999$.

\subsection{Topic Generation and Cluster Inference}

After model training, learned $G$ and $E$ will build a two-way projection between document-topic distribution and document-word distribution. Thus, $G$ and $E$ could be used for topic generation and cluster inference. 

To generate the word distribution of each topic, we use $\vec ts_{(k)}$, a $K$-dimensional vector, as the one-hot encoding of the $k$-th topic.  For example, $\vec ts_{2}=[0,1,0,0,0,0]^{\textrm T}$ in a six topic setting.  And the word distribution of the $k$-th topic is obtained by:
\begin{equation}
\vec \phi_{k}=G(\vec ts_{(k)})
\end{equation} 
Likewise,  given the document representation $\vec d_{r}$, topic distribution $\vec \theta_{r}$ obtained by BAT/Gaussian-BAT could be used for cluster inference based on:
\begin{align}
\vec \theta_{r}=E(\vec d_{r}); \quad 
c_{r}=\arg \max \vec\theta_{r}
\end{align}
where $c_{r}$ denotes the inferred cluster of $\vec d_{r}$.

\section{Experiments}

In this section, we first present the experimental setup which includes the datasets used and the baselines, followed by the experimental results. 

 \begin{figure*}[!h]
\centering
\includegraphics[
  width=1.0\textwidth,
  keepaspectratio]
{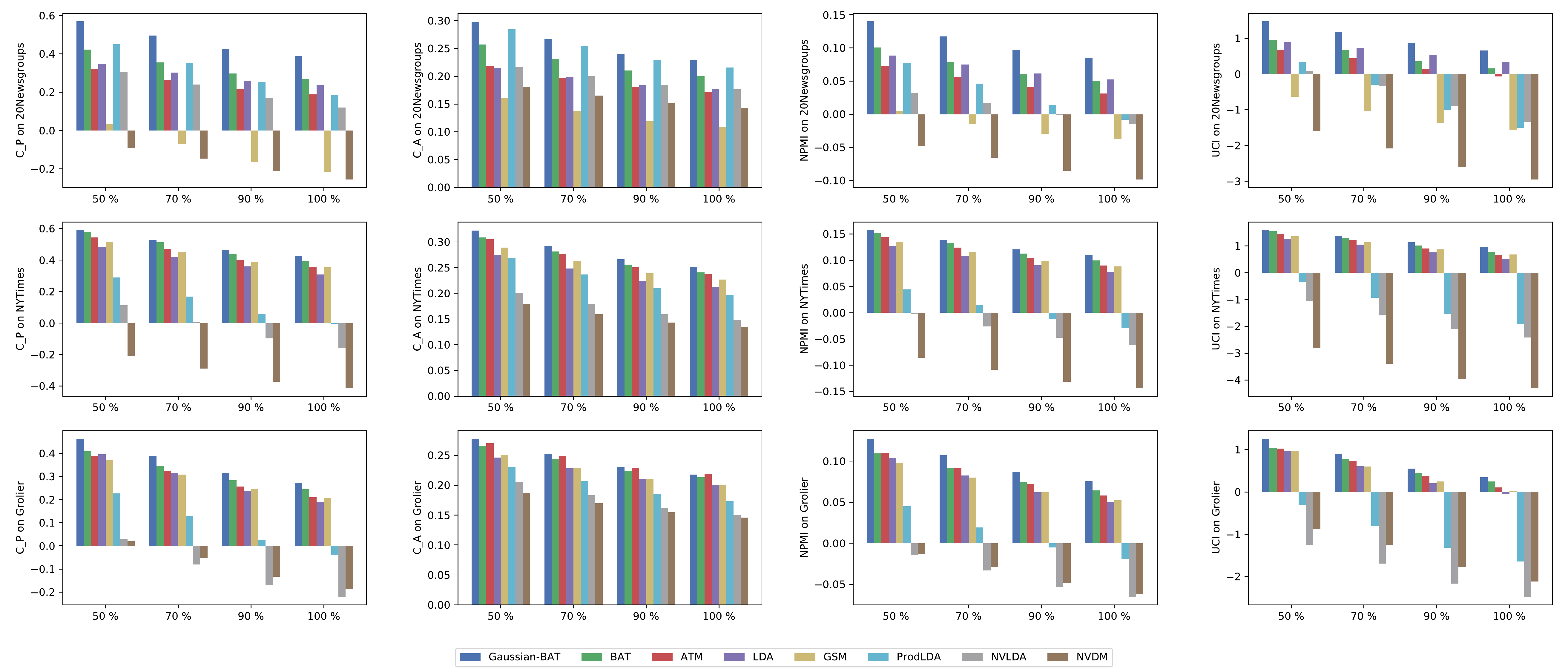}
%\caption{The comparison of average topic coherence vs. different topic proportion on 20Newsgroups, Grolier and NYTimes.}

\caption{The comparison of average topic coherence vs. different topic proportion on three datasets.}
\label{fig:cmp_bar}
\end{figure*}

\subsection{Experimental Setup}

We evaluate BAT and Gaussian-BAT on three datasets  for topic extraction and text clustering, 20Newsgroups\footnote{http://qwone.com/~jason/20Newsgroups/ \label{20news}}, Grolier\footnote{https://cs.nyu.edu/$\sim$roweis/data/ \label{grolier}} and NYTimes\footnote{http://archive.ics.uci.edu/ml/datasets/Bag+of+Words \label{nytimes}}.  Details are summarized below:

\noindent\emph{\underline{20Newsgroups}}~\cite{Lang95} is a collection of approximately 20,000 newsgroup articles, partitioned evenly across 20 different newsgroups.  \\
\noindent\underline{\emph{Grolier}} is built from Grolier Multimedia Encycopedia, which covers almost all the fields in the world.\\
\noindent\underline{\emph{NYTimes}} is a collection of news articles published between 1987 and 2007, and contains a wide range of topics, such as sports, politics, education, etc.

We use the full datasets of 20Newsgroups$^{\ref{20news}}$ and Grolier$^{\ref{grolier}}$. For the NYTimes dataset, we randomly select 100,000 articles and remove the low frequency words. The final statistics are shown in Table~\ref{tbs:statistic}:

\begin{table}[h]
%\vspace{-10pt}
\centering
\small
\scalebox{0.98}{
\begin{tabular}{lccc}
\hline
{\bfseries Dataset}& {\bfseries \#Doc (Train)}&{\bfseries \#Doc (Test)}&{\bfseries \#Words} \\
\hline
20Newsgroups&11,259 &7,488&1,995\\
Grolier& 29,762&-& 15,276\\
NYtimes  & 99,992&-&12,604\\
\hline
\end{tabular}
}
\caption{The statistics of datasets.}
\label{tbs:statistic}
\end{table}

We choose the following models as baselines:

\noindent\underline{\textbf{LDA}}~\cite{blei2003latent} extracts topics based on word co-occurrence patterns from documents. We implement LDA following the parameter setting suggested in~\cite{griffiths2004finding}.\\
\noindent\underline{\textbf{NVDM}}~\cite{miao2016neural} is an unsupervised text modeling approach based on VAE. We use the original implementation of the paper\footnote{https://github.com/ysmiao/nvdm}.\\
{\color{black}\noindent\underline{\textbf{GSM}}\cite{miao2017discovering} is an enhanced topic model based on NVDM, we use the original implementation in our experiments\footnote{https://github.com/linkstrife/NVDM-GSM}. }\\
\noindent\underline{\textbf{NVLDA}}~\cite{srivastava2017autoencoding}, also built on VAE but with the logistic-normal prior. We use the implementation provided by the author\footnote{https://github.com/akashgit/autoencoding vi for topic models\label{vae_lda}}.\\
\noindent\underline{\textbf{ProdLDA}}~\cite{srivastava2017autoencoding}, is a variant of NVLDA, in which the distribution over individual words is a product of experts. The original implementation is used. \\
{\color{black}\noindent\underline{\textbf{ATM}}~\cite{wang2019atm}, is a neural topic modeling approach based on adversarial training, we implement the ATM following the parameter setting suggested in the original paper. 
}

\subsection{Topic Coherence Evaluation}

Topic models are typically evaluated with the likelihood of held-out documents and topic coherence. However, Chang et al.~\shortcite{chang2009reading} showed that a higher likelihood of held-out documents does not correspond to human judgment of topic coherence. Thus, we follow~\cite{roder2015exploring} and employ four topic coherence metrics (C\_P, C\_A, NPMI and UCI) to evaluate the topics generated by various models.
%: C\_P (a metric based on sliding window, a one-preceding segmentation of the given words), C\_A (a metric based on a context window, a pairwise comparison of the given words and an indirect confirmation measure that uses normalized pointwise mutual information and the cosine similarity), NPMI (a metric based on the normalized pointwise mutual information) and UCI (a metric based on a sliding window and pointwise mutual information of all word pairs of given topics).
 In all experiments, each topic is represented by the top 10 words according to the topic-word probabilities, and all the topic coherence values are calculated using the Palmetto library\footnote{https://github.com/dice-group/Palmetto}.

\begin{table}[!h]
\centering
%\small
%\scriptsize
\footnotesize
\scalebox{0.76}{
\begin{tabular}{c|l|ccccc}

\hline
{\bfseries Dataset}&{\bfseries Model}& {\bfseries C\_P}&{\bfseries C\_A}&{\bfseries NPMI}&\bfseries UCI\\
\hline
\multirow{8}{*}{20Newsgroups}&NVDM&-0.2558 &0.1286 &-0.0984 &-2.9496\\
&GSM& -0.2318 & 0.1067 & -0.0400 &-1.6083\\
&NVLDA&0.1205 &0.1763  &-0.0207 &-1.3466\\
&ProdLDA&0.1858 &0.2155 &-0.0083 &-1.5044\\
&LDA&0.2361 &0.1769  &0.0523 &0.3399 \\
&ATM&0.1914 &0.1720  &0.0207 &-0.3871\\
&BAT&0.2597 &0.1976 &0.0472 &0.0969\\
&Gaussian-BAT&{\bfseries 0.3758} &{\bfseries 0.2251} &{\bfseries 0.0819} &{\bfseries 0.5925}\\
\hline
\multirow{8}{*}{Grolier}&NVDM&-0.1877 &0.1456 &-0.0619 &-2.1149\\
&GSM& 0.1974& 0.1966 & 0.0491 &-0.0410\\
&NVLDA&-0.2205 &0.1504  &-0.0653 &-2.4797\\
&ProdLDA&-0.0374 &0.1733 &-0.0193 &-1.6398\\
&LDA&0.1908 &0.2009  &0.0497&-0.0503 \\
&ATM& 0.2105& {\bfseries 0.2188} &0.0582 &0.1051\\
&BAT&0.2312 &0.2108 &0.0608 &0.1709\\
&Gaussian-BAT&{\bfseries 0.2606}&0.2142 &{\bfseries 0.0724} &{\bfseries 0.2836} \\
\hline
\multirow{8}{*}{NYtimes}&NVDM&-0.4130 &0.1341  &-0.1437 &-4.3072 \\
&GSM& 0.3426 & 0.2232 & 0.0848 &0.6224\\
&NVLDA&-0.1575 &0.1482  &-0.0614 &-2.4208 \\
&ProdLDA&-0.0034 &0.1963  &-0.0282 &-1.9173\\
&LDA&0.3083 &0.2127 & 0.0772 &0.5165 \\
&ATM& 0.3568& 0.2375 &0.0899 &0.6582\\
&BAT&0.3749 &0.2355 &0.0951 &0.7073\\
&Gaussian-BAT&{\bfseries 0.4163} &{\bfseries 0.2479} &{\bfseries 0.1079} &{\bfseries 0.9215}\\
\hline

\end{tabular}}
%\label{table:topic_cmp}
%\caption{Average topic coherence on 20Newsgroups, Grolier and NYtimes with five topic settings [20, 30, 50, 75, 100].}

\caption{Average topic coherence on three datasets with five topic settings [20, 30, 50, 75, 100].}
\label{tbs:average}
\end{table}

\begin{figure*}[h]
	\centering
	\includegraphics[
	width=1.0\textwidth]
	{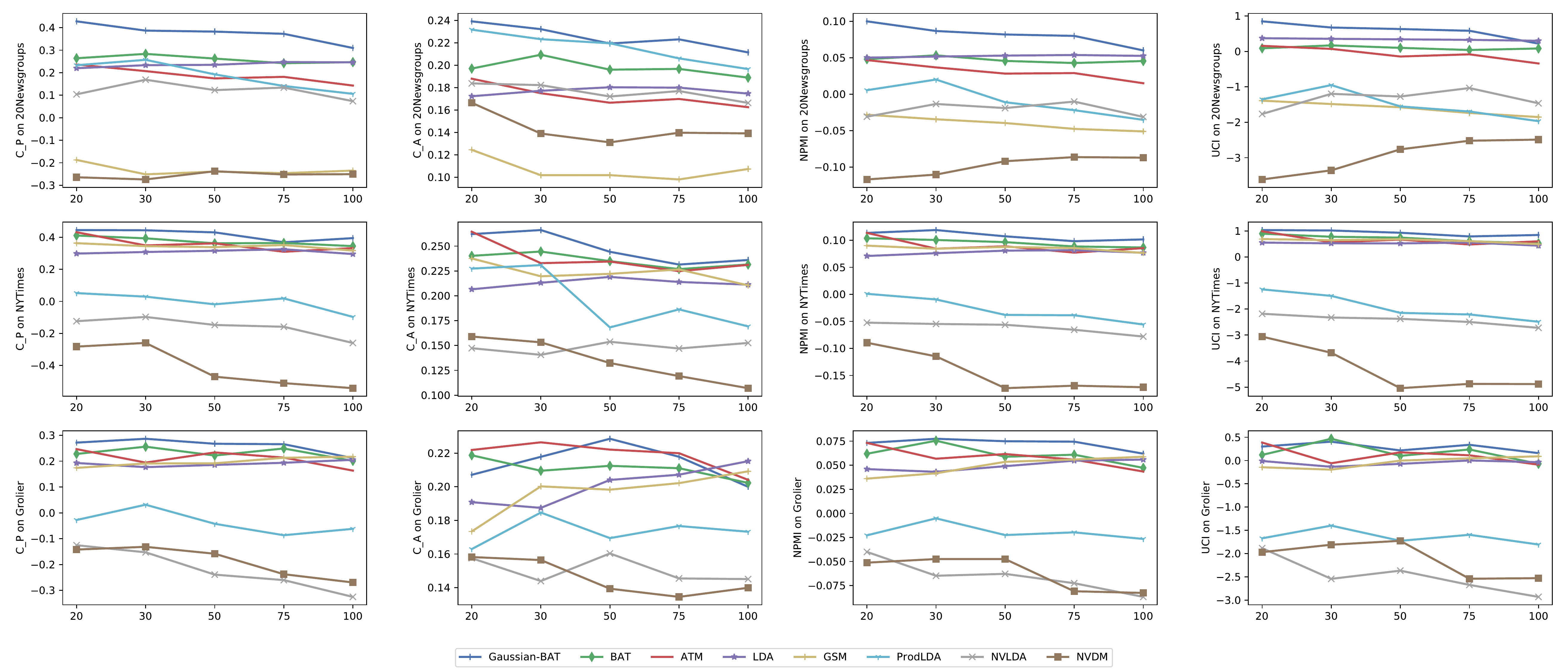}
	\caption{The comparison of average topic coherence vs. different topic number on 20Newsgroups, Grolier and NYTimes.}
	\label{fig:cmp_curve}
\end{figure*}

\begin{table*}[ht]
\centering
\small
%\scriptsize
%\footnotesize
{\color{black}
\scalebox{0.95}{
\begin{tabular}{c|l}
\hline
{\bfseries Model}& \multicolumn{1}{c}{\bfseries Topics}\\
\hline
\multirow{4}{*}{Gaussian-BAT}& voter campaign poll candidates democratic election republican vote presidential democrat \\
&song album music band rock pop sound singer jazz guitar  \\
&film movie actor character movies director series actress young scenes  \\
& flight airline passenger airlines aircraft shuttle airport pilot carrier planes \\
%& stock market fund investor investment index growth trading portfolio economy \\
\hline
\multirow{4}{*}{BAT}& vote president voter campaign election democratic governor republican black candidates \\
& album band music rock song jazz guitar pop musician \emph{record} \\
& film actor play acting role playing character father movie actress \\
& flight airline delay airlines plane pilot airport passenger carrier attendant \\
%& stock fund million shares investor investment mutual \emph{percent} initial offering \\
\hline
\multirow{4}{*}{LDA} & voter vote poll election campaign primary candidates republican race party\\
& music song band sound \emph{record} artist album show musical rock\\
& film movie character play actor director movies \emph{minutes} theater cast\\
& flight plane \emph{ship} \emph{crew} air pilot \emph{hour} \emph{boat} passenger airport\\
%& stock market \emph{percent} investor analyst \emph{quarter} investment shares share fund\\
\hline
\multirow{4}{*}{{\color{black}ATM}}& voter vote poll republican race \emph{primary percent} election campaign democratic\\
& music song musical album jazz band \emph{record recording} mp3 composer\\
& film movie actor director award movies character \emph{theater production} play\\
& jet flight airline \emph{hour} plane passenger \emph{trip plan travel} pilot\\
%& brokerage securities broker lender buyer transaction investor investment stock borrower\\
\hline
\end{tabular}
}
}
\label{table:stu}
\caption{Topic examples {\color{black}extracted by models}, italics means out-of-topic words. These topics correspond to `election', `music', `film' and `airline' respectively, and topic examples of other models are omitted due to poor quality.}
\label{tbs:example_topics}
\end{table*}

%To compare the performance of the proposed BAT and Gaussian-BAT with baselines comprehensively, 
We firstly make a comparison of topic coherence vs. different topic proportions. Experiments are conducted on the datasets with five topic number settings [20, 30, 50, 75, 100]. We calculate the average topic coherence values among topics whose coherence values are ranked at the top 50$\%$, 70$\%$, 90$\%$, 100$\%$ positions. For example, to calculate the average C\_P value of BAT $@90\%$, {\color{black}we first compute the average C\_P coherence with the selected topics whose C\_P values are ranked at the top 90\% for each topic number setting, and then average the five coherence values with each corresponding to a particular topic number setting.}

{\color{black}The detailed comparison is shown in Figure~\ref{fig:cmp_bar}}. {\color{black}It can be observed that BAT outperforms the baselines on all the coherence metrics for NYTimes datasets. For Grolier dataset, BAT outperforms all the baselines on C\_P, NPMI and UCI metrics, but gives slightly worse results compared to ATM on C\_A. For 20Newsgroups dataset, BAT performs the best on C\_P and NPMI, but gives slightly worse results compared to ProdLDA on C\_A, and LDA on UCI. 
By incorporating word embeddings through trainable Gaussian distribution, Gaussian-BAT outperforms all the baselines and BAT on four coherence metrics, often by a large margin, across all the three datasets except for Grolier dataset on C\_A when considering 100\% topics. This may be attribute to the following factors: (1) The Dirichlet prior employed in BAT and Gaussian-BAT could exhibit a multi-modal distribution in latent space and is more suitable for discovering semantic patterns from text; (2) ATM does not consider the relationship between topic distribution and word distribution since it only carry out adversarial training in word distribution space;  (3) The incorporation of word embeddings in Gaussian-BAT helps generating more coherent topics.}

We also compare the average topic coherence values (all topics taken into account) numerically to show the effectiveness of proposed BAT and Gaussian-BAT. The results of numerical topic coherence comparison are listed in Table~\ref{tbs:average} and each value is calculated by averaging the average topic coherences over five topic number settings. 
The best coherence value on each metric is highlighted in bold. It can be observed that Gaussian-BAT gives the best overall results across all metrics and on all the datasets except for Grolier dataset on C\_A. {\color{black} To make the comparison of topics more intuitive, we provide four topic examples extracted by models in Table~\ref{tbs:example_topics}. {\color{black}It can be observed that the proposed BAT and Gaussian-BAT can generate more coherent topics.} 
%we also provide the comparison of average topic coherence vs. different topic numbers in \emph{supplementary material}.

}

{\color{black}Moreover, to explore how topic coherence varies with different topic numbers,} we also provide the comparison of average topic coherence vs. different topic number on 20newsgroups, Grolier and NYTimes (all topics taken into account). The detailed comparison is shown in Figure~\ref{fig:cmp_curve}. {\color{black}It could be observed that Gaussian-BAT outperforms the baselines with 20, 30, 50 and 75 topics except for Grolier dataset on C\_A metric. However, when the topic number is set to 100, Gaussian-BAT performs slightly worse than LDA (e.g., UCI for 20Newsgroups and C\_A for NYTimes). This may be caused by the increased model complexity due to the larger topic number settings. Likewise, BAT can achieve at least the second-best results among all the approaches in most cases for NYTimes dataset. For Grolier, BAT also performs the second-best except on C\_A metric.   However, for 20newsgroups, the results obtained by BAT are worse than ProdLDA (C\_A) and LDA (UCI) due to the limited training documents in the dataset, though it still largely outperforms other baselines.} 

\subsection{Text Clustering}

We further compare our proposed models with baselines on text clustering. 
%To compare the performance of text clustering, the standard evaluation metric is employed. 
Due to the lack of document label information in Grolier and NYTimes, we only use 20Newsgroups dataset in our experiments. The topic number is set to 20 (ground-truth categories) and the performance is evaluated by accuracy $(ACC)$:
\begin{equation}
ACC=\max_{\rm map} \frac{\sum_{i=1}^{N_{t}}{\rm ind}(l_{i}={\rm map}(c_{i}))}{N_{t}}
\end{equation}
where $N_{t}$ is the number of documents in the test set, ${\rm ind(\cdot)}$ is the indicator function, $l_{i}$ is the ground-truth label of $i$-th document, $c_{i}$ is the category assignment, {\color{black}and ${\rm map}$ ranges over all possible one-to-one mappings between labels and clusters}. {\color{black} The optimal map function can be obtained by the Kuhn-Munkres algorithm~\cite{kuhn1955hungarian}. A larger accuracy value indicates a better text clustering results. }
\begin{table}[h]
\centering
\small
\scalebox{0.85}{
\begin{tabular}{lccccc}
\hline
{\bfseries Dataset}& {\bfseries NVLDA}&{\bfseries ProdLDA}&{\bfseries LDA}&{\bfseries BAT}&{\bfseries G-BAT} \\
\hline
20NG& 33.31\%& 33.82\%& 35.36\%& 35.66\%& {\bfseries 41.25}\%\\
\hline
\end{tabular}}
\caption{Text clustering accuracy on 20Newsgroups (20NG). `G-BAT' refers to `Gaussian-BAT'. The best result is highlighted in bold.}
\label{tbs:unsupervised_classification}
\end{table}  

The comparison of text clustering results on 20Newsgroups is shown in Table~\ref{tbs:unsupervised_classification}. Due to the poor performance of NVDM in topic coherence evaluation, its result is excluded here. {\color{black}Not surprisingly, NVLDA and ProdLDA perform worse than BAT and Gaussian-BAT that model topics with the Dirichlet prior. This might be caused by the fact that Logistic-Normal prior does not exhibit multiple peaks at the vertices of the simplex, as depicted in Figure~\ref{fig:distributions}. Compared with LDA, BAT achieves a comparable result in accuracy since both models have the same Dirichlet prior assumption over topics and only employ the word co-occurrence information.  Gaussian-BAT outperforms the second best model, BAT, by nearly 6\% in accuracy. This shows that the incorporation of word embeddings is important to improve the semantic coherence of topics and thus results in better consistency between cluster assignments and ground-truth labels. }

\section{Conclusion}
In this paper, we have explored the use of bidirectional adversarial training in neural topic models and proposed two novel approaches: the Bidirectional Adversarial Topic (BAT) model and the Bidirectional Adversarial Topic model with Gaussian (Gaussian-BAT). BAT models topics with the Dirichlet prior and builds a two-way transformation between document-topic distribution and document-word distribution via bidirectional adversarial training. Gaussian-BAT extends from BAT by incorporating word embeddings into the modeling process, thereby naturally considers the word relatedness information captured in word embeddings. The experimental comparison on three widely used benchmark text corpus with the existing neural topic models shows that BAT and Gaussian-BAT achieve improved topic coherence results. {\color{black}In the future, we would like to devise a nonparametric neural topic model based on adversarial training. Besides, developing correlated topic modelsis another promising direction.}

\section*{Acknowledgements}

We would like to thank anonymous reviewers for their valuable comments and helpful suggestions. This work was funded by the National Key Research and Development Program of China(2017YFB1002801) and the National Natural Science Foundation of China (61772132). And YH is partially supported by EPSRC (grant no. EP/T017112/1).

\bibliography{acl2020}
\bibliographystyle{acl_natbib}

\end{document}